\documentclass{article}

    \usepackage[final, nonatbib]{neurips_2019}

\usepackage[utf8]{inputenc} 
\usepackage[T1]{fontenc}    
\usepackage{hyperref}       
\usepackage{url}            
\usepackage{booktabs}       
\usepackage{amsfonts}       
\usepackage{nicefrac}       
\usepackage{microtype}      
\usepackage{graphicx}
\usepackage{amsmath}
\usepackage{amsfonts}
\usepackage{color}
\usepackage{subcaption}
\usepackage{footmisc}
\usepackage{bm}
\usepackage{cite}
\usepackage[noend]{algpseudocode}
\usepackage{algorithmicx,algorithm}
\title{AttentionXML: Label Tree-based Attention-Aware Deep Model for High-Performance Extreme Multi-Label Text Classification}

\author{
    Ronghui You$^1$, Zihan Zhang$^1$, Ziye Wang$^2$, Suyang Dai$^1$, \\
    \textbf{Hiroshi Mamitsuka$^{4,5}$, Shanfeng Zhu$^{1,3,*}$} \\
    $^1$ Shanghai Key Lab of Intelligent Information Processing, School of Computer Science, \\
    $^2$ Centre for Computational Systems Biology, School of Mathematical Sciences, \\
    $^3$ Shanghai Institute of Artificial Intelligence Algorithms and ISTBI, \\
    Fudan University, Shanghai, China; \\
    $^4$ Bioinformatics Center, Institute for Chemical Research, Kyoto University, Japan; \\
    $^5$ Department of Computer Science, Aalto University, Espoo and Helsinki, Finland \\
    \texttt{\{rhyou18,zhangzh17,zywang17,sydai16\}@fudan.edu.cn} \\
    \texttt{mami@kuicr.kyoto-u.ac.jp, zhusf@fudan.edu.cn}
}

\begin{document}

\maketitle

\begin{abstract}
Extreme multi-label text classification (XMTC) is an important problem in the 
era of {\it big data}, for tagging a given text with the most relevant multiple 
labels from an extremely large-scale label set. XMTC can be found in many 
applications, such as item categorization, web page tagging, and news 
annotation.
Traditionally most methods used bag-of-words (BOW) as inputs, ignoring word 
context as well as deep semantic information. Recent attempts to overcome the 
problems of BOW by deep learning still suffer from 1) failing to capture the 
important subtext for each label and 2) lack of scalability against the huge 
number of labels.
We propose a new label tree-based deep learning model for XMTC, called 
AttentionXML, with two unique features: 1) a multi-label attention mechanism 
with raw text as input, which allows to capture the most relevant part of text 
to each label; and 2) a shallow and wide probabilistic label tree (PLT), which 
allows to handle millions of labels, especially for "tail labels".
We empirically compared the performance of AttentionXML with those of eight 
state-of-the-art methods over six benchmark datasets, including Amazon-3M with 
around 3 million labels. AttentionXML outperformed all competing methods 
under all experimental settings.
Experimental results also show that AttentionXML achieved the best performance 
against tail labels among label tree-based methods. The code and datasets are 
available at \url{http://github.com/yourh/AttentionXML} .

\end{abstract}

\section{Introduction}

Extreme multi-label text classification (XMTC) is a natural language processing
(NLP) task for tagging each given text with its most relevant multiple labels
from an extremely large-scale label set. XMTC predicts multiple labels for a 
text, which is different from multi-class classification, where each instance has 
only one associated label. Recently, XMTC has become increasingly important, due 
to the fast growth of the data scale. In fact, over hundreds of thousands, even 
millions of labels and samples can be found in various domains, such as item 
categorization in e-commerce, web page tagging, news annotation, to name a few. 
XMTC poses great computational challenges for developing effective and efficient
classifiers with limited computing resource, such as an extremely large number of 
samples/labels and a large number of "tail labels" with very few positive 
samples.

Many methods have been proposed for addressing the challenges of XMTC. They can
be categorized into the following four types: 
1) 1-vs-All \cite{babbar2017dismec, yen2016pd, yen2017ppdsparse, babbar2019data},
2) Embedding-based \cite{bhatia2015sparse, tagami2017annexml}, 
3) Instance \cite{prabhu2014fastxml, jain2016extreme} or label tree-based \cite{jasinska2016extreme, prabhu2018parabel, wydmuch2018no, khandagale2019bonsai}) and
4) Deep learning-based methods \cite{liu2017deep} 
(see Appendix for more descriptions on these methods). The most related methods 
to our work are deep learning-based and label tree-based methods. 
A pioneering deep learning-based method is XML-CNN \cite{liu2017deep},
which uses a convolutional neural network (CNN) and dynamic pooling to learn the text representation. 
XML-CNN however cannot capture the most relevant parts of the input text to each label, 
because the same text representation is given for all labels. Another type of deep 
learning-based methods is sequence-to-sequence (Seq2Seq) learning-based 
methods, such as MLC2Seq\cite{nam2017maximizing}, SGM\cite{yang2018sgm} and 
SU4MLC\cite{lin2018semantic}. These Seq2Seq learning-based methods use a 
recurrent neural network (RNN) to encode a given raw text and an attentive RNN as a 
decoder to generate predicted labels sequentially. 
However the underlying assumption of these models is not reasonable since in 
reality there are no orders among labels in multi-label classification. In addition, 
the requirement of extensive computing in the existing deep learning-based methods makes 
it unbearable to deal with datasets with millions of labels. 

To handle such extreme-scale datasets, label tree-based methods use a
probabilistic label tree (PLT) \cite{jasinska2016extreme} to partition labels,
where each leaf in PLT corresponds to an original label and each internal node 
corresponds to a pseudo-label (meta-label). Then by maximizing a lower bound 
approximation of the log likelihood, each linear binary classifier for a tree 
node can be trained independently with only a small number of relevant 
samples\cite{prabhu2018parabel}.
Parabel \cite{prabhu2018parabel} 
is a state-of-the-art label tree-based method using bag-of-words (BOW) features. This
method constructs a binary balanced label tree by recursively partitioning nodes into two balanced clusters
until the cluster size (the number of labels in each cluster) is less than a given value (e.g. 100).
This produces a "deep" tree (with a high tree depth) for an extreme scale dataset, which deteriorates 
the performance due to an inaccurate approximation of likelihood, and the 
accumulated and propagated errors along the tree.
In addition, using balanced clustering with a large cluster size, many tail 
labels are combined with other dissimilar labels and grouped into one cluster. 
This reduces the classification performance on tail labels. 
On the other hand, another PLT-based 
method EXTREMETEXT\cite{wydmuch2018no}, which is based on 
FASTTEXT\cite{joulin2017bag}, uses dense features instead of BOW. Note that EXTREMETEXT ignores the order of words without 
considering context information, which underperforms Parabel. 

We propose a label tree-based deep learning model, AttentionXML, to address the 
current challenges of XMTC. AttentionXML uses raw text as its features with 
richer semantic context information than BOW features. AttentionXML is expected 
to achieve a high accuracy by using a BiLSTM (bidirectional long short-term
memory) to capture long-distance dependency among words and a multi-label 
attention to capture the most relevant parts of texts to each label.
Most state-of-the-art methods, such as DiSMEC\cite{babbar2017dismec} and 
Parabel\cite{prabhu2018parabel}, used only one representation for all 
labels including many dissimilar (unrelated) tail labels. It is difficult 
to satisfy so many dissimilar labels by the same representation. With 
multi-label attention, AttentionXML represents a given text differently 
for each label, which is especially helpful for many tail labels.
In addition, by using a shallow and wide PLT and a top-down level-wise model 
training, AttentionXML can handle extreme-scale datasets. 
Most recently, Bonsai\cite{khandagale2019bonsai} also uses shallow and diverse 
PLTs by removing the balance constraint in the tree construction, which improves
the performance by Parabel. Bonsai, however, needs high space complexity, such 
as a 1TB memory for extreme-scale datasets, because of using linear classifiers.
Note that we conceive our idea that is independent from Bonsai, and apply it in 
deep learning based method using deep semantic features other than BOW features 
used in Bonsai. The experimental results over six benchmarks datasets including 
Amazon-3M\cite{mcauley2013hidden} with around 3 million labels and 2 millions 
samples show that AttentionXML outperformed other state-of-the-art methods with 
competitive costs on time and space. 
The experimental results also show that AttentionXML is the best label tree-based method against tail labels.

\section{AttentionXML}

\begin{figure}[t]
\centering
\begin{subfigure}{0.52\textwidth}
\centering
\begin{subfigure}{\textwidth}
\centering
\includegraphics[width=\textwidth,trim=0 0 0 -3.5cm,clip]{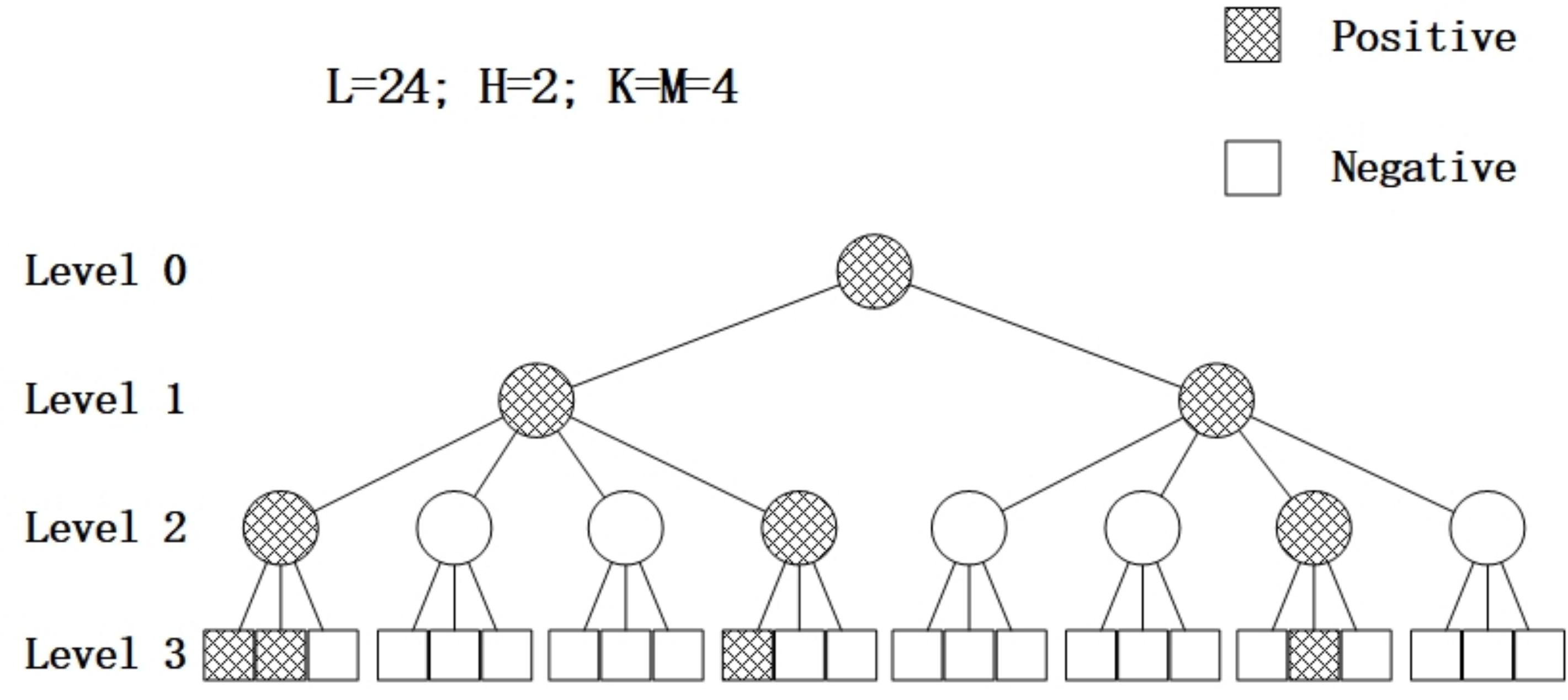}
\caption{}
\label{fig:tree}
\end{subfigure}
\begin{subfigure}{\textwidth}
\centering
\includegraphics[width=0.9\textwidth,trim=0 0 0 0 clip]{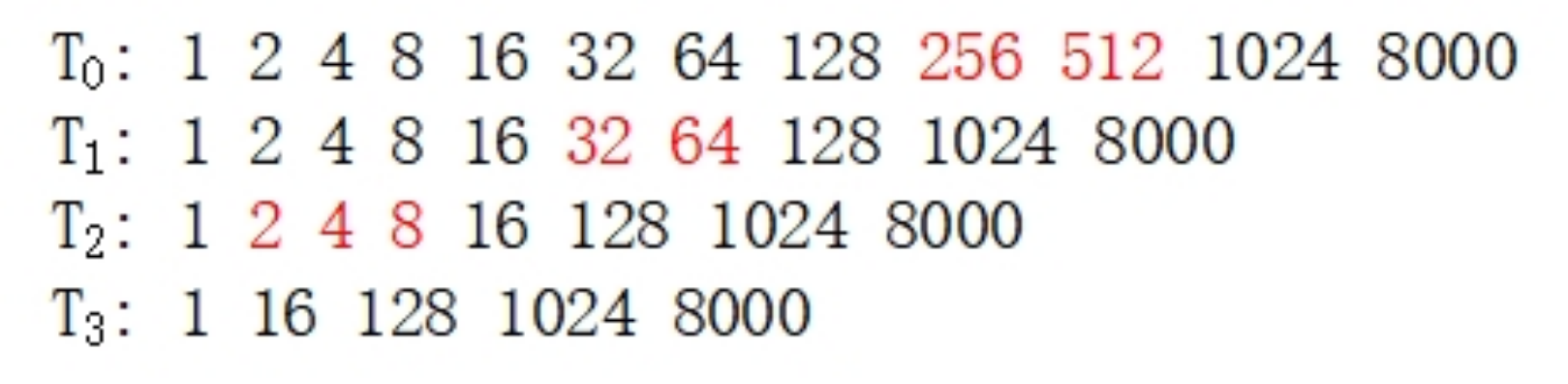}
\caption{}
\label{fig:example}
\end{subfigure}

\end{subfigure}
\begin{subfigure}{0.47\textwidth}
\centering
\includegraphics[width=\textwidth,trim=0 0.2cm 0 0.2cm,clip]{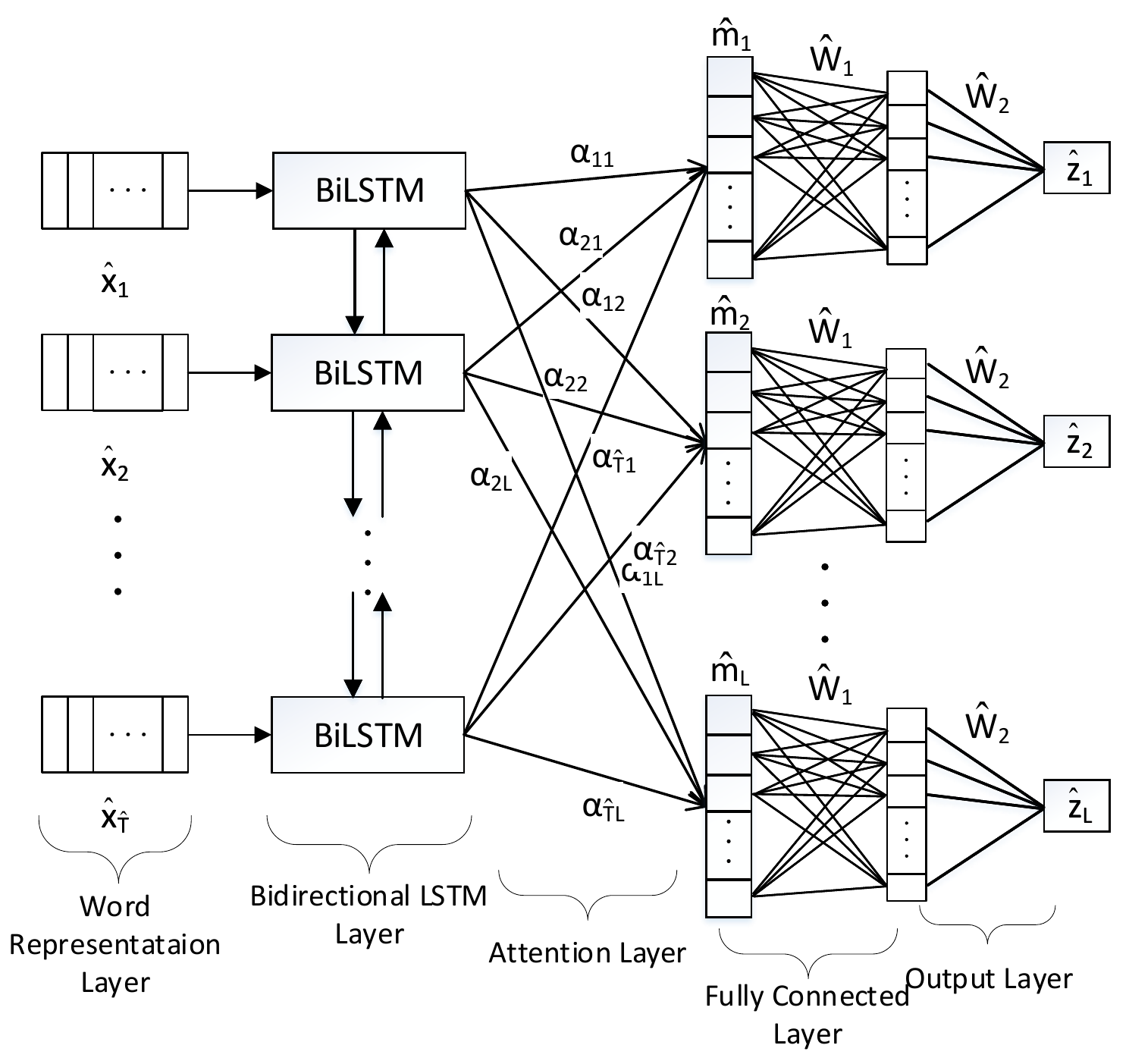}
\caption{}
\label{fig:deep}
\end{subfigure}
\caption{Label tree-based deep model AttentionXML for XMTC. 
(a) An example of PLT used in AttentionXML. 
(b) An example of a PLT building process with settings of $K=M=8=2^3$ and $H=3$ for $L=8000$. The numbers from left to right show those of nodes for each level from top to down. The numbers in red show those of nodes in $T_h$ that are removed in order to obtain $T_{h+1}$.
(c) Overview of attention-aware deep model in AttentionXML with text (length $\hat{T}$) as its input and predicted scores $\hat{z}$ as its output. The $\hat{x}_i\in\mathbb{R}^{\hat{D}}$ is the embeddings of $i$-th wo rd(where $\hat{D}$ is the dimension of embeddings), $\alpha\in\mathbb{R}^{\hat{T}\times L}$ are the attention coefficients and $\hat{W}_1$ and $\hat{W}_2$ are parameters of the fully connected layer and output layer. }
\label{fig:overview}
\end{figure}

\subsection{Overview}
The main steps of AttentionXML are: (1) building a shallow and wide PLT (Figs. \ref{fig:tree} and \ref{fig:example});
and (2) for each level $d~(d>0)$ of a given constructed PLT, training an 
attention-aware deep model AttentionXML$_d$ with a BiLSTM and a multi-label 
attention (Fig. \ref{fig:deep}). The pseudocodes for constructing PLT, 
training and prediction of AttentionXML are presented in \textbf{Appendix}.

\subsection{Building Shallow and Wide PLT}
PLT\cite{jain2016extreme} is a tree with $L$ leaves where each leaf corresponds to an original label. Given a sample $x$,
we assign a label $z_{n}\in \{0, 1\}$ for each node $n$, which 
indicates whether the subtree rooted at node $n$ has a leaf (original label) 
relevant to this sample. PLT estimates the conditional probability 
$P(z_n|z_{Pa(n)}=1, x)$ to each node $n$. The marginal probability $P(z_n=1|x)$
for each node $n$ can be easily derived as follows by the chain rule of 
probability:
\begin{equation}
P(z_n=1|x) = \prod_{i\in Path(n)} P(z_i=1|z_{Pa(i)}=1, x)
\end{equation}
where $Pa(n)$ is the parent of node $n$ and $Path(n)$ is the set of nodes on the
path from node $n$ to the root (excluding the root).

As mentioned in \textbf{Introduction}, large tree height $H$ (excluding the root
and leaves) and large cluster size $M$ will harm the performance. So in 
AttentionXML, we build a shallow (a small $H$) and wide (a small $M$) PLT $T_H$.
First, we built an initial PLT, $T_0$, by a top-down hierarchical clustering, 
which was used in Parabel\cite{prabhu2018parabel}, with a small cluster size 
$M$. In more detail, we represent each label by normalizing the sum of BOW 
features of text annotated by this label. The labels are then recursively 
partitioned into two smaller clusters, which correspond to internal tree nodes, 
by a balanced $k$-means ($k$=2) until the number of labels smaller than $M$
\cite{prabhu2018parabel}.
$T_0$ is then compressed into a shallow and wide PLT, i.e. $T_H$, which is a 
$K (=2^c)$ ways tree with the height of $H$. 
This compress operation is similar to the pruning strategy in some hierarchical multi-class classification methods\cite{babbar2016learning, babbar2013flat}.
We first choose all parents of 
leaves as $\bm{S}_0$ and then conduct compress operations $H$ times, resulting 
in $T_H$. The compress operation has three steps: for example in the $h$-th 
compress operation over $T_{h-1}$, we (1) choose $c$-th ancestor nodes ($h<H$) 
or the root ($h=H$) as $\bm{S}_h$, (2) remove nodes between $\bm{S}_{h-1}$ and 
$\bm{S}_h$, and (3) then reset nodes in $\bm{S}_{h}$ as parents of corresponding
nodes in $\bm{S}_{h-1}$. This finally results in a shallow and wide tree $T_H$. 
Practically we use $M=K$ so that each internal node except the root has no more
than $K$ children.
Fig \ref{fig:example} shows an example of building PLT. More examples can be found in \textbf{Appendix}.

\subsection{Learning AttentionXML}
Given a built PLT, training a deep model against nodes at a deeper level is more
difficult because nodes at a deeper level have less positive examples. Training 
a deep model for all nodes of different levels together is hard to optimize and 
harms the performance, which can only speed up marginally. Thus we train 
AttentionXML in a level-wise manner as follows:

\begin{enumerate}
\item 
AttentionXML trains a single deep model for each level of a given PLT in a 
top-down manner. Note that labeling each level of the PLT is still a 
multi-label classification problem. For the nodes of first level (children of
the root), AttentionXML (named AttentionXML$_1$ for the first level) can be 
trained for these nodes directly.

\item
AttentionXML$_d$ for the $d$-th level ($d>1$) of the given PLT is only trained 
by candidates $g(x)$ for each sample $x$. Specifically, we sort nodes of the 
($d-1$)-th level by $z_n$ (from positives to negatives) and then their scores 
predicted by AttentionXML$_{d-1}$ in the descending order. We keep the top $C$ 
nodes at the ($d-1$)-th level and choose their children as $g(x)$. 
It's like a kind of additional negative sampling and we can get a more precise 
approximation of log likelihood than only using nodes with positive parents.

\item 
During prediction, for the $i$-th sample, the predicted score $\hat{y}_{ij}$
for $j$-th label can be computed easily based on the probability chain 
rule. For the prediction efficiency, we use beam search  
\cite{prabhu2018parabel,khandagale2019bonsai}: for the $d$-th ($d>1$) level we 
only predict scores of nodes that belong to nodes of the ($d-1$)-th level with
top $C$ predicted scores.
\end{enumerate}

We can see that the deep model without using a PLT can be regarded as a special 
case of AttentionXML with a PLT with only the root and $L$ leaves.

\subsection{Attention-Aware Deep Model}

Attention-aware deep model in AttentionXML consists of five layers:
1) Word Representation Layer,
2) Bidirectional LSTM Layer,
3) Multi-label Attention Layer,
4) Fully Connected Layer and 5) Output Layer.
Fig. \ref{fig:deep} shows a schematic picture of attention-aware deep model in AttentionXML.

\subsubsection{Word Representation Layer}
The input of AttentionXML is raw tokenized text with length $\hat{T}$. Each word is represented by a 
deep semantic dense vector, called {\it word embedding} 
\cite{pennington2014glove}. 
In our experiments, we use pre-trained 300-dimensional GloVe 
\cite{pennington2014glove} word embedding as our initial word representation.

\subsubsection{Bidirectional LSTM Layer}
RNN is a type of neural network with a memory state to process sequence inputs. 
Traditional RNN has a problem called {\it gradient vanishing and exploding} 
during training \cite{bengio1994learning}. Long short-term memory 
(LSTM) \cite{hochreiter1997long} is proposed for solving this problem.
We use a Bidirectional LSTM (BiLSTM) to capture both the left- and right-sides 
context (Fig. \ref{fig:deep}), where at each time step $t$ the output 
$\mathbf{\hat{h}}_t$ is obtained by concatenating the forward output 
$\overrightarrow{\mathbf{h}}_t$ and the backward output 
$\overleftarrow{\mathbf{h}}_t$. 

\subsubsection{Multi-Label Attention}
Recently, an {\it attention mechanism} in neural networks has been 
successfully used in many NLP tasks, such as machine translation, machine 
comprehension, relation extraction, and speech 
recognition\cite{bahdanau2014neural, luong2015effective}.
The most relevant context to each label can be different in XMTC. AttentionXML 
computes the (linear) combination of context vectors $\mathbf{\hat{h}}_i$ for 
each label through a {\it multi-label attention mechanism}, inspired by 
\cite{lin2017structured}, to capture various intensive parts of a text. 
That is, the output of multi-label attention layer 
$\mathbf{\hat{m}}_j\in\mathbb{R}^{2\hat{N}}$ of the $j$-th label can be obtained as follows:
\begin{equation}
\mathbf{\hat{m}}_j = \sum_{i=1}^{\hat{T}} \alpha_{ij}\mathbf{\hat{h}}_i,
\qquad
\alpha_{ij} = \frac{e^{\mathbf{\hat{h}}_i\mathbf{\hat{w}}_j}}{\sum_{t=1}^{\hat{T}} e^{\mathbf{\hat{h}}_{t}\mathbf{\hat{w}}_j}},
\end{equation}
where $\alpha_{ij}$ is the normalized coefficient of $\mathbf{\hat{h}}_i$ and
$\mathbf{\hat{w}}_j\in\mathbb{R}^{2\hat{N}}$ is the so-called attention 
parameters. Note that $\hat{w}_j$ is different for each label.

\subsubsection{Fully Connected and Output Layer}
AttentionXML has one (or two) fully connected layers and one output layer. 
The same parameter values are used for all labels at the fully connected (and output) layers, 
to emphasize differences of attention among all labels. 
Also sharing 
the parameter values of fully connected layers among all labels can  
largely reduce the number of parameters to avoid overfitting and keep the model 
scale small.

\subsubsection{Loss Function}
AttentionXML uses the binary cross-entropy loss function, which is used in XML-CNN 
\cite{liu2017deep} as the loss function. Since the number of labels for each 
instance varies, we do not normalize the predicted probability which is done in multi-class 
classification. 

\subsection{Initialization on parameters of AttentionXML}
We initialize the parameters of AttentionXML$_d$ ($d>1$) by using the parameters
of trained AttentionXML$_{d-1}$, except the attention layers. This 
initialization helps models of deeper levels converge quickly, resulting in 
improvement of the final accuracy.

\subsection{Complexity Analysis}
The deep model without a PLT is hard to deal with extreme-scale datasets,
because of high time and space complexities of the multi-label attention 
mechanism. Multi-label attention in the deep model needs $O(BL\hat{N}\hat{T})$ time 
and $O(BL(\hat{N}+\hat{T}))$ space for each batch iteration, where $B$ is the 
batch size. 
For large number of labels ($L$ > 100k), 
the time cost is huge.
Also the whole model cannot be saved in the 
limited memory space of GPUs.
On the other hand, the time complexity of AttentionXML with a PLT is much smaller than that 
without a PLT, although we need train $H+1$ different deep models. That is, the 
label size of AttentionXML$_1$ is only $L/K^H$, which is much smaller than $L$. 
Also the number of candidate labels of  AttentionXML$_{d} (d>1)$ is only 
$C\times K$, which is again much smaller than $L$. Thus our efficient 
label tree-based AttentionXML can be run even with the limited GPU memory.

\section{Experimental Results}

\subsection{Dataset}
We used six most common XMTC benchmark datasets (Table \ref{tab:dataset}):
three large-scale datasets ($L$ ranges from ~4K to 30K) :
EUR-Lex\footnote{\url{http://www.ke.tu-darmstadt.de/resources/eurlex/eurlex.html}} \cite{mencia2008efficient}, 
Wiki10-31K\footnote{\url{http://manikvarma.org/downloads/XC/XMLRepository.html}\label{XMLData}} \cite{zubiaga2012enhancing},
and AmazonCat-13K \footref{XMLData} \cite{mcauley2013hidden};
and three extreme-scale datasets ($L$ ranges from 500K to 3M):
Amazon-670K\footref{XMLData}\cite{mcauley2013hidden}, 
Wiki-500K\footref{XMLData}
and Amazon-3M\footref{XMLData}\cite{mcauley2013hidden}.
Note that both Wiki-500K and Amazon-3M have around two million samples for training.

\begin{table*}[t]
\vspace{-2mm}
\small
\centering
\caption{Datasets we used in our experiments.}
\label{tab:dataset}
\begin{tabular}{@{}crrrrrrrrrrrrrrrrrr@{}}
\hline
Dataset & $N_{train}$ & $N_{test}$ & $D$ & $L$ & $\overline{L}$ & $\hat{L}$ & $\overline{W}_{train}$ & $\overline{W}_{test}$ \\
\hline
EUR-Lex & 15,449 & 3,865 & 186,104 & 3,956 & 5.30 & 20.79 & 1248.58 & 1230.40 \\
Wiki10-31K & 14,146 & 6,616 & 101,938 & 30,938 & 18.64 & 8.52 & 2484.30 & 2425.45 \\
AmazonCat-13K & 1,186,239 & 306,782 & 203,882 & 13,330 & 5.04 & 448.57 & 246.61 & 245.98 \\
Amazon-670K & 490,449 & 153,025 & 135,909 & 670,091 & 5.45 & 3.99 & 247.33 & 241.22 \\
Wiki-500K & 1,779,881 & 769,421 & 2,381,304 & 501,008 & 4.75 & 16.86 & 808.66 & 808.56 \\
Amazon-3M & 1,717,899 & 742,507 & 337,067 & 2,812,281 & 36.04 & 22.02 & 104.08 & 104.18 \\
\hline	
\end{tabular}

 $N_{train}$: \#training instances,
 $N_{test}$: \#test instances, 
 $D$: \#features, 
 $L$: \#labels,
 $\overline{L}$: average \#labels per instance,
 $\hat{L}$: the average \#instances per label, 
 $\overline{W}_{train}$: the average \#words per training instance and $\overline{W}_{test}$: the average \#words per test instance. 
 The partition of training and test is from the data source.
\vspace{-3mm}
\end{table*}

\subsection{Evaluation Measures}
We chose P@$k$ (Precision at $k$) \cite{jain2016extreme} as our evaluation 
metrics for performance comparison, since $P@k$ is widely used for evaluating 
the methods for XMTC.
\begin{equation}
P@k = \frac{1}{k} \sum_{l=1}^k \mathbf{y}_{rank(l)} \\
\end{equation}
where $\mathbf{y} \in \{0, 1\}^L$ is the true binary vector, and $rank(l)$ is 
the index of the $l$-th highest predicted label. Another common evaluation 
metric is $N@k$ (normalized Discounted Cumulative Gain at $k$). Note that $P@1$ is equivalent to $N@1$. We evaluated performance by $N@k$, and confirmed that the performance of $N@k$ kept the same trend as $P@k$.
We thus omit the results of $N@k$ in the
main text due to space limitation (see \textbf{Appendix}).

\subsection{Competing Methods and Experimental Settings}

We compared the state-of-the-art and most representative XMTC methods 
(implemented by the original authors) with AttentionXML:
AnnexML\footnote{\url{https://s.yimg.jp/dl/docs/research_lab/annexml-0.0.1.zip}} (embedding),
DiSMEC\footnote{\url{https://sites.google.com/site/rohitbabbar/dismec}} (1-vs-All),
MLC2Seq\footnote{\url{https://github.com/JinseokNam/mlc2seq.git}} (deep learning),
XML-CNN\footref{XMLData} (deep learning),
PfastreXML\footref{XMLData} (instance tree),
Parabel\footref{XMLData} (label tree) and
XT\footnote{\url{https://github.com/mwydmuch/extremeText}} (ExtremeText) (label tree) and
Bonsai\footnote{\url{https://github.com/xmc-aalto/bonsai}} (label tree).

For each dataset, we used the most frequent words in the training set as a 
limited-size vocabulary (not over 500,000). Word embeddings were fine-tuned 
during training except EUR-Lex and Wiki10-31K. 
We truncated each text after 500 words for training and predicting efficiently. 
We used dropout\cite{srivastava2014dropout} to avoid overfitting after the 
embedding layer with the drop rate of 0.2 and after the BiLSTM with the drop 
rate of 0.5. Our model was trained by Adam\cite{kingma2014adam} with the 
learning rate of 1e-3. We also used SWA (stochastic weight 
averaging)\cite{izmailov2018averaging} with a constant learning rate to enhance 
the performance. 
We used a three PLTs ensemble in AttentionXML similar to Parabel\cite{prabhu2018extreme} 
and Bonsai\cite{khandagale2019bonsai}.
We also examined performance of AttentionXML with only one PLT (without ensemble), 
called AttentionXML-1.
On three large-scale datasets, we used AttentionXML with a PLT including only a 
root and $L$ leaves(which can also be considered as the deep model without PLTs).
Other hyperparameters in our experiments are shown in Tabel \ref{tab:resource}.

\begin{table}
\centering
\small
\caption{Hyperparameters we used in our experiments, practical computation time and model size.}
\label{tab:resource}
\begin{tabular}{@{}c|ccccccc|ccccccccccccccc@{}}
\hline
Datasets & $E$ & $B$ & $\hat{N}$ & $\hat{N}_{fc}$ & $H$ & $M$ & $C$ & Train & Test & Model Size \\
&&&&&&$=K$&& (hours) & (ms/ & (GB) \\
&&&&&&&&& sample) \\
\hline
EUR-Lex         & 30 & 40 & 256 & 256 & - & - & - &0.51 & 2.07 & 0.20 \\
Wiki10-31K      & 30 & 40 & 256 & 256 & - & - & - & 1.27 & 4.53 & 0.62 \\
AmazonCat-13K   & 10 & 200 & 512 & 512,256 & - & - & -  & 13.11 & 1.63 & 0.63 \\
Amazon-670K		& 10 & 200 & 512 & 512,256 & 3 & 8 & 160 & 13.90 & 5.27 & 5.52  \\
Wiki-500K		& 5 & 200 & 512 & 512,256 & 1 & 64 & 15 & 19.55 & 2.46 & 3.11  \\
Amazon-3M		& 5 & 200 & 512 & 512,256 & 3 & 8 & 160 & 31.67 & 5.92 & 16.14  \\
\hline
\end{tabular}

$E$: The number of epoch;
$B$: The batch size;
$N$: The hidden unit size of LSTM;
$N_{fc}$: The hidden unit size of fully connected layers;
$H$: The height of PLT (excluding the root and leaves);
$M$: The maximum cluster size; 
$K$: The parameters of the compress process, and here we set $M=K=2^c$;
$C$: The number of parents of candidate nodes.
\end{table}

\subsection{Performance comparison}
\begin{table}
\caption{Performance comparisons of AttentionXML and other competing methods over six benchmarks. The results with the stars are from \textbf{Extreme Classification Repository} directly. }
\label{tab:per:pk}
\begin{subtable}[t]{0.49\textwidth}
	\centering
	\begin{tabular}{@{}cccccccccccccccccccccccccccccc@{}}
		\hline
		Methods & P@1=N@1 & P@3 & P@5 \\
		\hline\hline
		\multicolumn{4}{c}{EUR-Lex} \\
		\hline 
		AnnexML         & 79.66 & 64.94 & 53.52 \\
		DiSMEC          & 83.21 & 70.39 & 58.73 \\
		PfastreXML      & 73.14 & 60.16 & 50.54 \\
		Parabel         & 82.12 & 68.91 & 57.89 \\
		XT              & 79.17 & 66.80 & 56.09 \\
		Bonsai          & 82.30 & 69.55 & 58.35 \\
		MLC2Seq         & 62.77 & 59.06 & 51.32 \\
		XML-CNN         & 75.32 & 60.14 & 49.21 \\
		AttentionXML-1  & \it{85.49} & \it{73.08} & \it{61.10} \\
		AttentionXML    & \textbf{87.12} & \textbf{73.99} & \textbf{61.92} \\
		\hline\hline
		\multicolumn{4}{c}{Wiki10-31K} \\
		\hline
		AnnexML         & 86.46 & 74.28 & 64.20 \\
		DiSMEC          & 84.13 & 74.72 & 65.94 \\
		PfastreXML*     & 83.57 & 68.61 & 59.10 \\
		Parabel         & 84.19 & 72.46 & 63.37 \\
		XT              & 83.66 & 73.28 & 64.51 \\
		Bonsai          & 84.52 & 73.76 & 64.69 \\
		MLC2Seq         & 80.79 & 58.59 & 54.66 \\
		XML-CNN         & 81.41 & 66.23 & 56.11 \\
		AttentionXML-1  & \it{87.05} & \it{77.78} & \it{68.78} \\
		AttentionXML    & \textbf{87.47} & \textbf{78.48} & \textbf{69.37} \\
		\hline\hline
		\multicolumn{4}{c}{AmazonCat-13K} \\
		\hline
		AnnexML         & 93.54 & 78.36 & 63.30 \\
		DiSMEC          & 93.81 & 79.08 & 64.06 \\
		PfastreXML*     & 91.75 & 77.97 & 63.68 \\
		Parabel         & 93.02 & 79.14 & 64.51 \\
		XT              & 92.50 & 78.12 & 63.51 \\
		Bonsai          & 92.98 & 79.13 & 64.46 \\
		MLC2Seq         & 94.29 & 69.45 & 57.55 \\
		XML-CNN         & 93.26 & 77.06 & 61.40 \\
		AttentionXML-1  & \it{95.65} & \it{81.93} & \it{66.90}\\
		AttentionXML    & \textbf{95.92} & \textbf{82.41} & \textbf{67.31} \\
		\hline
	\end{tabular}
\end{subtable}
\quad
\begin{subtable}[t]{0.49\textwidth}
	\centering
	\begin{tabular}{@{}cccccccccccccccccccccccccccccc@{}}
		\hline
		Methods & P@1=N@1 & P@3 & P@5 \\
		\hline\hline
		\multicolumn{4}{c}{Amazon-670K} \\
		\hline 
		AnnexML         & 42.09 & 36.61 & 32.75 \\
        DiSMEC			& 44.78 & 39.72 & 36.17 \\
		PfastreXML*     & 36.84 & 34.23 & 32.09 \\
        Parabel      	& 44.91 & 39.77 & 35.98 \\
        XT      	    & 42.54 & 37.93 & 34.63 \\
        Bonsai  		& 45.58 & 40.39 & 36.60 \\
        MCL2Seq         & - & - & - \\
        XML-CNN 		& 33.41 & 30.00 & 27.42 \\
        AttentionXML-1  & \it{45.66} & \it{40.67} & \it{36.94} \\
        AttentionXML    & \textbf{47.58} & \textbf{42.61} & \textbf{38.92} \\
		\hline\hline
        \multicolumn{4}{c}{Wiki-500K} \\
        \hline
        AnnexML			& 64.22 & 43.15 & 32.79 \\
        DiSMEC          & 70.21 & 50.57 & 39.68 \\
        PfastreXML		& 56.25 & 37.32 & 28.16 \\
        Parabel         & 68.70 & 49.57 & 38.64 \\
        XT          	& 65.17 & 46.32 & 36.15 \\
        Bonsai  		& 69.26 & 49.80 & 38.83 \\
        MCL2Seq         & - & - & - \\
        XML-CNN         & - & - & - \\
        AttentionXML-1  & \it{75.07} & \it{56.49} & \it{44.41} \\
        AttentionXML    & \textbf{76.95} & \textbf{58.42} & \textbf{46.14} \\
        \hline\hline
        \multicolumn{4}{c}{Amazon-3M} \\
        \hline
        AnnexML 		& \it{49.30} & 45.55 & 43.11 \\
        DiSMEC* 		& 47.34 & 44.96 & 42.80 \\
        PfastreXML*	    & 43.83 & 41.81 & 40.09 \\
        Parabel         & 47.42 & 44.66 & 42.55 \\
        XT      		& 42.20 & 39.28 & 37.24 \\
        Bonsai  		& 48.45 & 45.65 & 43.49 \\
        MCL2Seq         & - & - & - \\
        XML-CNN         & - & - & - \\
        AttentionXML-1  & 49.08 & \it{46.04} & \it{43.88} \\
        AttentionXML    & \textbf{50.86} & \textbf{48.04} & \textbf{45.83} \\
        \hline
	\end{tabular}
\end{subtable}
\end{table}

Tables \ref{tab:per:pk} shows the performance results of AttentionXML and other competing 
methods by $P@k$ over all six benchmark datasets.
Following the previous work on XMTC, we focus on top predictions by varying $k$ at 1, 3 and 5 in $P@k$, resulting in 18 (= three k $\times$ six datasets) values of $P@k$ for each method.

1) AttentionXML (with a three PLTs ensemble) outperformed all eight 
competing methods by $P@k$. For example, for $P@5$, among all datasets, 
AttentionXML is at least $4\%$ higher than the second best method (Parabel 
on AmazonCat-13K).  For Wiki-500K, AttentionXML is even more than $17\%$ 
higher than the second best method (DiSMEC). AttentionXML also outperformed 
AttentionXML-1 (without ensemble), especially on three extreme-scale datasets.
That's because on extreme-scale datasets, the ensemble with different PLTs 
reduces more variance, while on large-scale datasets models the ensemble is 
with the same PLTs (only including the root and leaves). Note that AttentionXML-1 
is much more efficient than AttentionXML, because it only trains one model without 
ensemble.

2) AttentionXML-1 outperformed all eight competing methods by $P@k$, except 
only one case. Performance improvement was especially notable for EUR-Lex,
Wiki10-31K and Wiki-500K, with longer texts than other datasets (see Table \ref{tab:dataset}).
For example, for $P@5$, AttentionXML-1 achieved 44.41, 68.78 and 61.10 on 
Wiki-500K, Wiki10-31K and EUR-Lex, which were around $12\%$, $4\%$ and $4\%$ 
higher than the second best, DiSMEC with 39.68, 65.94 and 58.73, respectively. 
This result highlights that longer text has larger amount of context 
information, where multi-label attention can focus more on the most relevant parts
of text and extract the most important information on each label. 

3) Parabel, a method using PLTs, can be considered as taking the advantage of 
both tree-based (PfastreXML) and 1-vs-All (DiSMEC) methods. It 
outperformed PfastreXML and achieved a similar performance to DiSMEC (which is 
however much more inefficient). 
ExtremeText (XT) is an online learning method with PLTs (similar to Parabel), which used
dense instead of sparse representations and achieved slightly lower performance
than parabel.
Bonsai, another method using PLTs, outperformed 
Parabel on all datasets except AmazonCat-13K. In addition, Bonsai achieved 
better performance than DiSMEC on Amazon-670K and Amazon-3M. This result 
indicates that the shallow and diverse PLTs in Bonsai improves its 
performance. However, Bonsai needs much more memory than Parabel, for example, 
1TB memory for extreme-scale datasets.
Note that AttentionXML-1 with only one shallow and wide PLT, still significantly 
outperformed both Parabel and Bonsai on all extreme-scale datasets, especially 
Wiki-500K.

4) MLC2Seq, a deep learning-based method, obtained the worst performance on the 
three large-scale datasets, probably because of its unreasonable assumption. 
XML-CNN, another deep learning-based method with a simple dynamic pooling was 
much worse than the other competing methods, except MLC2Seq. Note that both 
MLC2Seq and XML-CNN are unable to deal with datasets with millions of labels. 

5) AttentionXML was the best method among all the competing methods, on the three
extreme-scale datasets (Amazon-670K, Wiki-500K and Amazon-3M).
Although the improvement by AttentionXML-1 over the second and third best methods (Bonsai 
and DiSMEC) is rather slight, AttentionXML-1 is much faster than DiSMEC and 
uses much less memory than Bonsai. In addition, the improvement by AttentionXML with a 
three PLTs ensemble over Bonsai and DiSMEC is more significant, which is still 
faster than DiSMEC and uses much less memory than Bonsai.

6) AnnexML, the state-of-the-art embedding-based method, reached the second best 
$P@1$ on Amazon-3M and Wiki10-31K, respectively. 
Note that the performance of AnnexML was not necessarily so on the other datasets.
The average number of labels per sample of Amazon-3M (36.04) and Wiki10-31K 
(18.64) is several times larger than those of other datasets (only around 5).
This suggests that each sample in these datasets has been well annotated. Under 
this case, embedding-based methods may acquire more complete information from 
the nearest samples by using KNN (k-nearset neighbors) and might gain a 
relatively good performance on such datasets.

\subsection{Performance on tail labels}

\begin{figure}
    \centering
    \includegraphics[width=1.0\textwidth]{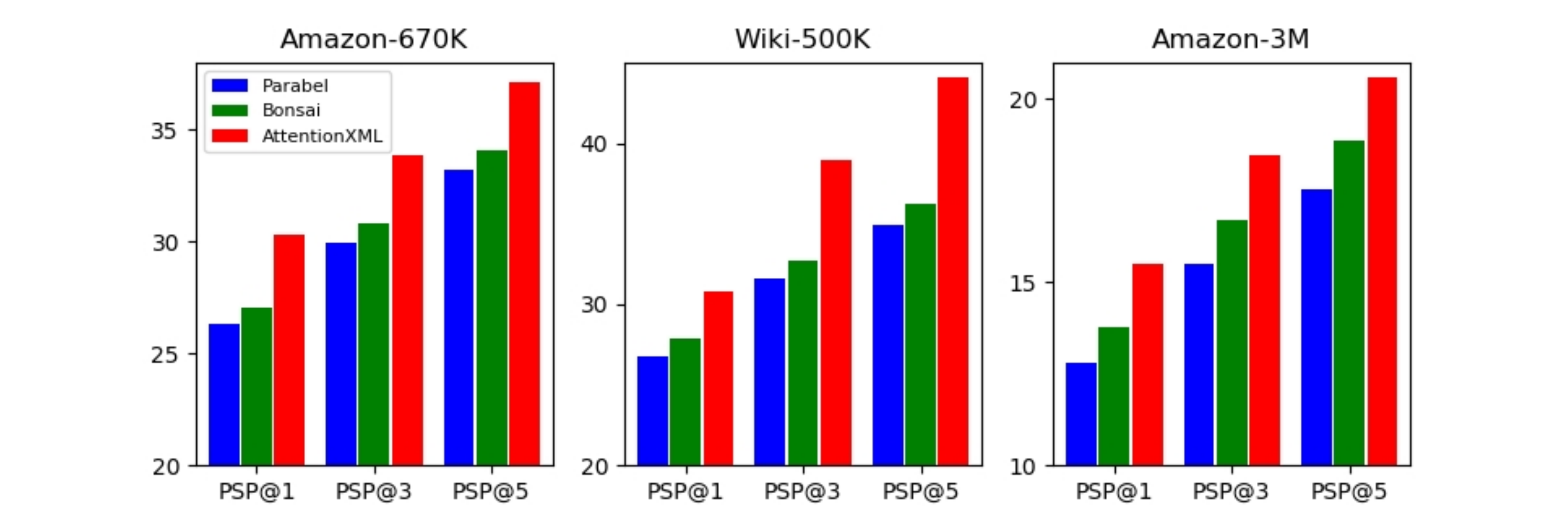}
    \caption{$PSP@k$ of label tree-based methods.}
    \label{fig:per:psp}
\end{figure}

We examined the performance on tail labels by $PSP@k$
(propensity scored precision at k)\cite{jain2016extreme}:
\begin{equation}
\text{PSP@}k = \frac{1}{k} \sum_{l=1}^k \frac{\mathbf{y}_{\text{rank}(l)}}{\mathbf{p}_{\text{rank}(l)}}
\end{equation}
where $\mathbf{p}_{\text{rank}(l)}$ is the propensity score\cite{jain2016extreme}
of label $rank(l)$.
Fig \ref{fig:per:psp} shows the results of three label tree-based 
methods (Parabel, Bonsai and AttentionXML) on the three extreme-scale datasets.
Due to space limitation, we reported $PSP@k$ results of AttentionXML and all compared
methods including ProXML\cite{babbar2019data} (a state-of-the-art method on $PSP@k$) 
on six benchmarks in \textbf{Appendix}.
 
AttentionXML outperformed both Parabel and Bonsai in $PSP@k$ on all datasets. AttentionXML use a shallow and wide PLT, which is different from Parabel. 
Thus this result indicates that this shallow and wide PLT in AttentionXML is 
promising to improve the performance on tail labels.
Additionally, multi-label attention in AttentionXML would be also effective for 
tail labels, because of capturing the most important parts of text for each 
label, while Bonsai uses just the same BOW features for all labels.

\subsection{Ablation Analysis}
For examining the impact of BiLSTM and multi-label attention, we also run a 
model which consists of a BiLSTM, a max-pooling (instead of the attention layer of AttentionXML), and the fully connected layers (from XML-CNN).
Tabel \ref{tab:per:ablation} shows the $P@5$ results on three large-scale 
datasets. BiLSTM outperformed XML-CNN on all three datasets, probably because of
capturing the long-distance dependency among words. 
AttentionXML (BiLSTM+Attn) further outperformed XML-CNN and BiLSTM, especially 
on EUR-Lex and Wiki10-31K, which have long texts. Comparing with a simple 
dynamic pooling, obviously multi-label attention can extract the most important 
information to each label from long texts more easily. 
In addition, Table \ref{tab:per:ablation} shows that SWA has a favorable effect on improving prediction 
accuracy.

\begin{table}
\centering
\caption{P@5 of XML-CNN, BiLSTM and AttentionXML (all without ensemble)}
\label{tab:per:ablation}
\begin{tabular}{@{}ccccccccccccccccccccc@{}}
\hline
Dataset & XML-CNN & BiLSTM & AttentionXML & AttentionXML\\
& & & (BiLSTM+Att) & (BiLSTM+Att+SWA) \\
\hline
EUR-Lex         & 49.21 & 53.12 & 59.61 & \textbf{61.10} \\
Wiki10-31K      & 56.21 & 59.55 & 66.51 & \textbf{68.78} \\
AmazonCat-13K   & 61.40 & 63.57 & 66.29 & \textbf{66.90} \\
\hline
\end{tabular}
\end{table}

\subsection{Impact of Number of PLTs in AttentionXML}
We examined the performance of ensemble with different number of PLTs in AttentionXML.
Table \ref{tab:per:ensemble} shows the performance comparison of AttentionXML with
different number of label trees. We can see that more trees much improve the prediction 
accuracy. However, using more trees needs much more time for both training and prediction.
So its a trade-off between performance and time cost.

\begin{table}[]
    \centering
    \caption{Performance of variant number of trees in AttentionXML.}
    \label{tab:per:ensemble}
    \begin{tabular}{@{}c|ccc|ccc|ccccccccccccccccccccccc@{}}
    \hline
    & \multicolumn{3}{c|}{Amazon-670K} & \multicolumn{3}{c|}{Wiki-500K} & \multicolumn{3}{c}{Amazon-3M} \\
    \hline
    Trees & P@1 & P@3 & P@5 & P@1 & P@3 & P@5 & P@1 & P@3 & P@5\\
    \hline
    1 & 45.66 & 40.67 & 36.94 & 75.07 & 56.49 & 44.41 & 49.08 & 46.04 & 43.88 \\
    2 & 46.86 & 41.95 & 38.27 & 76.44 & 57.92 & 45.68 & 50.34 & 47.45 & 45.26 \\
    3 & 47.58 & 42.61 & 38.92 & 76.95 & 58.42 & 46.14 & 50.86 & 48.04 & 45.83 \\
    4 & \textbf{48.03} & \textbf{43.05} & \textbf{39.32} & \textbf{77.21} & \textbf{58.72} & \textbf{46.40} & \textbf{51.66} & \textbf{48.39} & \textbf{46.23} \\
    \hline
    \end{tabular}
\end{table}

\subsection{Computation Time and Model Size}
AttentionXML runs on 8 Nvidia GTX 1080Ti GPUs. Table \ref{tab:resource} shows 
the computation time for training (hours) and testing (milliseconds/per 
sample), as well as the model size (GB) of \textbf{AttentionXML with only one PLT} for 
each dataset. For the ensemble of several trees, AttentionXML can be trained and predicted 
on a single machine sequentially, or on a distributed settings simultaneously 
and efficiently (without any network communication).

\section{Conclusion}
We have proposed a new label tree-based deep learning model, AttentionXML, for XMTC, with two distinguished features: the multi-label attention mechanism, which allows to capture the important parts of texts most relevant to each label,
and a shallow and wide PLT, which allows to handle millions of labels efficiently and effectively.
We examined the predictive performance of AttentionXML, comparing with eight 
state-of-the-art methods over six benchmark datasets including three 
extreme-scale datasets.
AttentionXML outperformed all other competing methods over all six datasets, 
particularly datasets with long texts. Furthermore, AttentionXML revealed the 
performance advantage in predicting long tailed labels. 

\subsubsection*{Acknowledgments}
S. Z. is supported by National Natural Science Foundation of China (No. 61872094 and No. 61572139) and Shanghai Municipal Science and Technology Major Project (No. 2017SHZDZX01). R. Y., Z. Z., Z. W., S. Y. are supported by the 111 Project (NO. B18015), the key project of Shanghai Science \& Technology (No. 16JC1420402), Shanghai Municipal Science and Technology Major Project (No. 2018SHZDZX01) and ZJLab.  H.M. has been supported in part by JST ACCEL [grant number JPMJAC1503], MEXT Kakenhi [grant numbers 16H02868 and 19H04169], FiDiPro by Tekes (currently Business Finland) and AIPSE by Academy of Finland.


\bibliographystyle{abbrv}
\bibliography{AttentionXML}

\newpage
\appendix
\section{Examples of PLT in AttentionXML}
Here we show PLTs we used in AttentionXML for three extreme-scale datasets:
\begin{enumerate}
    \item 
    For Amazon-670K with the number of labels $L=670,091$, we used a setting 
    of $H=3$ and $K=M=2^3=8$, the number of nodes in each level of the PLT we 
    used from top to down is 
    1; 2,048($2^{11}$); 16,384($2^{14}$); 131,072($2^{17}$) and 670,091, 
    respectively.
    
    \item
    For Wiki-500K with the number of labels $L=501,008$, we used a setting 
    of $H=1$ and $K=M=2^6=64$, the number of nodes in each level of the PLT we 
    used from top to down is 
    1; 8,192($2^{13}$) and 501,008, 
    respectively.
    
    \item 
    For Amazon-3M with the number of labels $L=2,812,281$, we used a setting 
    of $H=3$ and $K=M=2^3=8$, the number of nodes in each level of the PLT we 
    used from top to down is 
    1; 8,192($2^{13}$); 65,536($2^{16}$); 524,288($2^{19}$) and 2,812,281, 
    respectively.
    
\end{enumerate}

\section{Algorithms}

\textbf{Algorithm \ref{algorithm:compress}} presents the pseudocode for compressing a 
deep PLT to a shallow one. The deep PLT can be generated by a hierarchical 
KMeans (K=2) following Parabel\cite{prabhu2018parabel}. \textbf{Algorithm 
\ref{algorithm:label}} presents the pseudocode for getting labels of tree nodes for each 
sample. \textbf{Algorithm \ref{algorithm:train}} and \textbf{Algorithm 
\ref{algorithm:predict}} presents the pseudocode of training and prediction of 
AttentionXML, respectively.

\begin{algorithm}
\caption{Compressing into a shallow and wide PLT}
\label{algorithm:compress}
\hspace*{0.02in} {\bf Input:}
(a) Labels of training data $\{y_i\}_{i=1}^{N_{train}}$;
(b) PLT $T_0$;
(c) $K=2^c$, $H$ \\
\hspace*{0.02in} {\bf Output:}
A compressed shallow and wide PLT $T$
\begin{algorithmic}[1]
\State $S_0 \gets \{$parent nodes of leaves$\}$
\For {$h \gets 1$ \textbf{to} $H$}
    \If {h < H}
        \State $S_h \gets \{c$-th ancestor node $n$ of nodes in $S_{h-1}\}$
    \Else
        \State $S_h \gets \{$ the root of $T_0\}$
    \EndIf
    \State $T_h \gets T_{h-1}$
    \ForAll {nodes $n \in S_h$}
        \ForAll {nodes $n' \in S_{h-1}$ and node $n$ is the ancestor of $n'$ in $T$}
            \State $Pa(n') \gets n$ 
            \Comment{Let node $n$ be parent of node $n'$ in $T_h$, $Pa(n)$ means parent of $n$.}
        \EndFor
    \EndFor
\EndFor
\State \Return $T_H$
\end{algorithmic}
\end{algorithm}

\begin{algorithm}
\caption{Getting labels of tree nodes}
\label{algorithm:label}
\hspace*{0.02in} {\bf Input:}
(a) Labels of training data $\{y_i\}_{i=1}^{N_{train}}$;
(b) PLT $T$ \\
\hspace*{0.02in} {\bf Output:}
Tree nodes labels $\{z_{i}\}_{i=1}^{N_{train}}$
\begin{algorithmic}[1]
\For {$i \gets 1$ \textbf{to} $N$}
    \ForAll {node $n$ in $T$}
        \State $z_{i,n} \gets 0$
    \EndFor
    \For {$j \gets 1$ \textbf{to} $L$}
        \If {$y_{i,j}$}
            \State $n \gets $ leaf node corresponding to label $j$ in $T$
            \While {$n$ isn't the root of $T$}
                \State $z_{i,n} \gets 1$
                \State $n \gets Pa(n)$
            \EndWhile
        \EndIf
    \EndFor
\EndFor
\State \Return $\{z_{i}\}_{i=1}^{N_{train}}$
\end{algorithmic}
\end{algorithm}

\begin{algorithm}
\caption{Training of AttentionXML}
\label{algorithm:train}
\hspace*{0.02in} {\bf Input:}
(a) Training data $\{x_i, z_i\}_{i=1}^{N_{train}}$;
(b) PLT $T$;
(c) Candidates number $C$;\\
\hspace*{0.02in} {\bf Output:} 
Trained Model AttentionXML$_1$, AttentionXML$_2$, ..., AttentionXML$_{H+1}$
\begin{algorithmic}[1]
\State $H \gets$ the height of $T$
\For {$i \gets 1$ to $N$}
    \State $c_i^1 \gets \{$ children node $n$ of the root of $T\}$
\EndFor
\For {$d \gets 1$ \textbf{to} $H+1$}
    \If {$d>1$}
        \For {$i \gets 1$ \textbf{to} $N$}
            \ForAll {node $n \in c_i^{d-1}$}
                \State $\hat{z}_{i,n} \gets$ score predicted by AttentionXML$_{d-1}$ with $x_i$
                \State $\hat{z}_{i,n} \gets \hat{z}_{i,n} \times \hat{z}_{i, Pa(n)}$
            \EndFor
            \State $s_i^d \gets $ Top $C$ nodes in $c_i^{d-1}$ by $z_i$ from positive to negative and $\hat{z}_i$ from large to small
            \State $c_i^d \gets \{$ All children nodes of $s_i^d$ in $T\}$
        \EndFor
        \State Initialize weights of AttentionXML$_{d}$ with weights of AttentionXML$_{d-1}$
    \EndIf
    \State Train AttentionXML$_d$ with $\{x_i, z_i, c_i^d\}_{i=1}^{N_{train}}$
\EndFor
\State \Return AttentionXML$_1$, AttentionXML$_2$, ..., AttentionXML$_{H+1}$
\end{algorithmic}
\end{algorithm}

\begin{algorithm}
\caption{Prediction of AttentionXML}
\label{algorithm:predict}
\hspace*{0.02in} {\bf Input:}
(a) Test sample $x$;
(b) PLT $T$;
(c) Candidates number $C$;\\
\hspace*{0.02in} {\bf Output:}
Ranked predicted labels 

\begin{algorithmic}[1]
\State $H \gets $ the height of $T$
\State $c^1 \gets \{$ All children nodes the root of $T\}$
\For {$d \gets 1$ \textbf{to} $H+1$}
    \If {$d>1$}
        \State $c^d \gets \{$ All children nodes of $s^{d-1}$ in $T\}$
    \EndIf
    \ForAll {node $n\in c^{d}$}
        \State $\hat{z}_n \gets$ score predicted by AttentionXML$_{d}$ with $x$
        \State $\hat{z}_n \gets \hat{z}_n \times \hat{z}_{Pa(n)}$
    \EndFor
    \State $s^d \gets$ Top $C$ nodes in $c^d$ by $\hat{z}$ from large to small
\EndFor
\State \Return Ranked labels corresponding to $s^{H+1}$
\end{algorithmic}
\end{algorithm}

\section{Related Work}
Existing work for XMTC can be categorized into the following four types: 
1) 1-vs-All, 2) Embedding-based, 3) Tree-based, and 4) Deep learning-based 
methods.

\subsection{1-vs-All Methods}
1-vs-All methods, such as 1-vs-All SVM, train a classifier (e.g. SVM) for each 
label independently. A clear weak point is that its computational complexity is 
very high, and the model size can be huge, due to the extremely large number of
labels and instances.

For reducing the complexity, PD-Sparse \cite{yen2016pd} and PPDSparse 
\cite{yen2017ppdsparse} are recently proposed by using the idea of sparse learning. PD-Sparse 
trains a classifier for each label by a margin-maximizing loss funcion with the 
$L_1$ penalty to obtain an extremely sparse solution both in primal and dual, 
without sacrificing the expressive power of the predictor.
PPDSparse \cite{yen2017ppdsparse} extends PD-Sparse, by using efficient 
parallelization of large-scale distributed computing (e.g. 100 cores), 
achieving a better performance than PD-Sparse.

As another state-of-the-art 1-vs-All method, DiSMEC \cite{babbar2017dismec} learns a linear classifier for each label
based on distributed computing.
DiSMEC uses a double layer of parallelization to sufficiently exploit computing 
resource (400 cores), implementing a significant speed-up of training and 
prediction. Pruning spurious weight coefficients (close to zero), DiSMEC makes the model thousands of times smaller, resulting in reducing the required
computational resource to a much smaller size than those by other 
state-of-the-art methods.

\subsection{Embedding-based Methods}
The idea of embedding-based methods is, since the label size is huge, to 
compress the labels and use the compressed labels for training, and finally, 
compressed labels  are decompressed for prediction. More specifically, given $n$
training instances $(\mathbf{x}_i, \mathbf{y}_i) (i=1,\dots,n)$, where 
$\mathbf{x}_i\in\mathbb{R}^d$ is a $d$-dimensional feature vector and 
$\mathbf{y}_i\in\{0,1\}^L$ is an L-dimensional label vector. Embedding-based 
methods compress $\mathbf{y}_i$ into a lower $\hat{L}$-dimensional embedding 
vector $\mathbf{z}_i$ by $\mathbf{z}_i = f_C(\mathbf{y}_i)$, where $f_C$ is 
called a compression function. Then embedding-based methods train regression 
model $f_R$ for predicting embedding vector $\mathbf{z}_i$ with input feature 
vector $\mathbf{x}_i$. For a given instance with feature vector $\mathbf{x}_i$, 
embedding-based methods predict its embedding vector $\mathbf{z}_i$ by 
$\mathbf{z}_i = f_R(\mathbf{x}_i)$ and predict label vector $\mathbf{\hat{y}}_i$
by $\mathbf{\hat{y}}_i = f_D(\mathbf{z}_i)$ where $f_D$ is called a 
decompression function. A disadvantage is that feature space $X$ and label space
$Y$ are projected into a low dimensional space $Z$ for efficiency. As such, some
information must be lost through this process, sometimes resulting in only 
limited success.

The main difference among embedding-based methods is the design of compression 
function $f_C$ and decompression function $f_D$. For example, the most 
representative method, SLEEC \cite{bhatia2015sparse}, learns embedding vectors 
$\mathbf{z}_i$ by capturing non-linear label correlations, preserving the 
pairwise distance between label vectors, $\mathbf{y}_i$ and $\mathbf{y}_j$,
i.e. $d(\mathbf{z}_i, \mathbf{z}_j) \approx d(\mathbf{y}_i, \mathbf{y}_j)$ if 
$i$ is in the $k$ nearest neighbors of $j$. Regressors $\mathbf{V}$ are then 
trained to predict embedding label $\mathbf{z}_i = \mathbf{V}\mathbf{x}_i$, and
a $k$-nearest neighbor classifier (KNN) is used for prediction. KNN has high 
computational complexity, so SLEEC uses clusters, into which training instances 
are embedded. That is, given a test instance, only the cluster into which this
instance can be fallen is used for prediction.

AnnexML \cite{tagami2017annexml} is an extension of SLEEC, solving the three 
problems of SLEEC: 1) clustering without labels; 2) ignoring the distance value 
in prediction (since just KNN is used); and 3) slow prediction. Addressing the above problems, AnnexML 
generates a KNN graph (KNNG) of label vectors in the embedding space, addressing
the above problems, and improves both accuracy and efficiency.

\subsection{Tree-based Methods}
Tree-based methods use the idea of (classical) decision tree. They generate a 
tree by recursively partitioning given instances by features at non-terminal
nodes, resulting in a simple classifier at each leaf with only a few active 
labels. Also following the idea of random forest, most tree-based methods 
generate an ensemble of trees, selecting (sampling) a feature subset randomly 
at each node of the trees. A clear disadvantage of the tree-based method is the low
performance, because selection at a node of each tree is just an approximation.

The most representative tree-based method, FastXML \cite{prabhu2014fastxml}, 
learns a hyperplane to split instances rather than to select a single feature.
In more detail, FastXML optimizes an nDCG (normalized Discounted Cumulative 
Gain)-based ranking loss function at each node. An extension of FastXML is 
PfastreXML \cite{jain2016extreme}, which keeps the same architecture as 
FastXML, and PfastreXML uses a propensity scored objective function, instead of
optimizing nDCG. Due to this objective function, PfastreXML makes more accurate
tail label prediction over FastXML.

Label tree-based methods are already described in the paper.

\subsection{Deep learning-based Methods}
Deep learning-based methods can be divided into two types: sequence-to-sequence (Seq2Seq) learning (S2SL) and discriminative learning-based (DL) methods.
As the DL methods are already explained in Introduction, we here focus on S2SL methods.
Pioneering approaches of S2SL are MLC2Seq \cite{nam2017maximizing}, SGM \cite{yang2018sgm},
and SU4MLC \cite{lin2018semantic}, all of which use 
an attention based Seq2Seq architecture \cite{bahdanau2014neural}. 
This architecture has the input with the representations of source text by an RNN encoder and predicts the labels with another attention based RNN decoder.
Also trainable attention parameters in this architecture are the same for all labels (Note that AttentionXML has label-specific attention parameters).
The difference from MLC2seq is that SGM considers the label distribution at the last time step in decoder,
and SU4MLC uses higher-level semantic unit representations 
by multi-level dilated convolution.
Empirically MLC2Seq is demonstrated to outperform FastXML (tree-based method) in terms of F1 
measure. In constrast, SGM and SU4MLC have shown no comparative performance advantages.

\section{Experiments and Results}

\subsection{Evaluation Metrics}
We chose $P@k$ (Precision at $k$) and $N@k$ (normalized Discounted 
Cumulative Gain at $k$) as our evaluation metrics for performance comparison, 
since both $P@k$ and $N@k$ are widely used for evaluation methods for 
multi-labelclassification problems. $P@k$ is defined as follows:
\begin{equation}
P@k = \frac{1}{k} \sum_{l=1}^k \mathbf{y}_{rank(l)}
\end{equation}
where $\mathbf{y} \in \{0, 1\}^L$ is the true binary vector, and $rank(l)$ is 
the index of the $l$-th highest predicted label. $N@k$ is defined as follows:
\begin{equation}
\begin{split}
DCG@k &= \sum^{k}_{l=1} \frac{\mathbf{y}_{rank(l)}}{log(l+1)} \\
iDCG@k &= \sum^{min(k, ||\mathbf{y}||_0)}_{l=1} \frac{1}{log(l+1)} \\
N@k &= \frac{DCG@k}{iDCG@k}
\end{split}
\end{equation}
$N@k$ is a metric for ranking, meaning that the order of top $k$ prediction
is considered in $N@k$ but not in $P@k$. Note that $P@1$ and $N@1$ are the 
same. We also use $PSP@k$(propensity scored precision at $k$) as our evaluation
metric for performance comparison on "tail labels"\cite{jain2016extreme}. 
$PSP@k$ is defined as follows:
\begin{equation}
PSP@k = \frac{1}{k} \sum_{l=1}^k \frac{\mathbf{y}_{\text{rank}(l)}}{\mathbf{p}_{\text{rank}(l)}}
\end{equation}
where $\mathbf{p}_{\text{rank}(l)}$ is the propensity score\cite{jain2016extreme} of label rank$(l)$.

\begin{figure}
\centering
\includegraphics[scale=0.75]{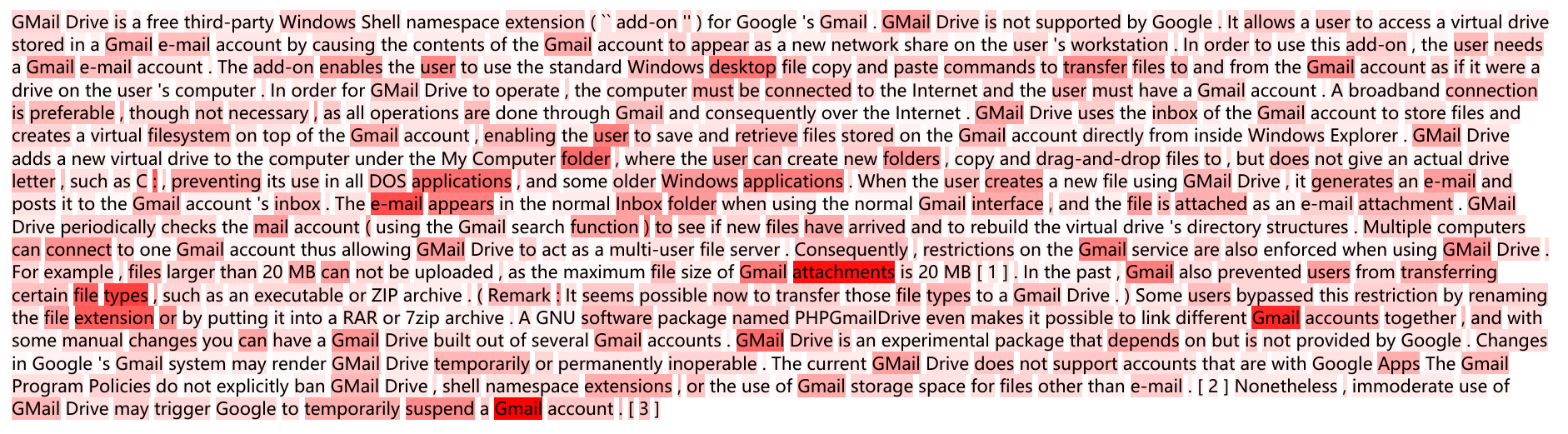}
\caption{Attention of a test instance (wiki entry: GMail Driver) to the label "gmail" in Wiki10-31K.}
\label{fig:attention}
\vspace{-3mm}
\end{figure}

\subsection{Performance Results}
Table \ref{tab:per} shows the performance comparisons of AttentionXML and other 
seven state-of-the-art methods over six benchmark datasets.

\begin{table*}
\centering
\caption{Performance comparisons of AttentionXML and other competing methods over six benchmark datasets.}
\label{tab:per}
\begin{tabular}{@{}ccccccccccccccccccccc@{}}
\hline
Methods & P@1=N@1 & P@3 & P@5 & N@3 & N@5 & PSP@1 & PSP@3 & PSP@5 \\
\hline\hline
\multicolumn{9}{c}{EUR-Lex} \\
\hline
AnnexML			& 79.66 & 64.94 & 53.52 & 68.70 & 62.71 & 33.88 & 40.29 & 43.69 \\
DiSMEC			& 83.21 & 70.38 & 58.73 & 73.73 & 67.96 & 38.45 & 46.20 & 50.25 \\
ProXML          & 83.41 & 70.97 & 58.94 & 74.23 & 68.16 & \it{44.92} & 48.37 & 50.75 \\
PfastreXML		& 73.13 & 60.16 & 50.54 & 63.51 & 58.71 & 41.68 & 44.01 & 45.73 \\
Parabel     	& 82.12 & 68.91 & 57.89 & 72.33 & 66.95 & 37.20 & 44.74 & 49.17 \\
Bonsai  		& 82.30 & 69.55 & 58.35 & 72.97 & 67.48 & 37.33 & 45.40 & 49.92 \\
XML-CNN 		& 75.32 & 60.14 & 49.21 & 63.95 & 58.11 & 32.41 & 36.95 & 39.45 \\
AttentionXML-1	& \it{85.49} & \it{73.08} & \it{61.10} & \it{76.37} & \it{70.49} & 44.75 & \it{51.29} & \it{53.86} \\
AttentionXML    & \textbf{87.12} & \textbf{73.99} & \textbf{61.92} & \textbf{77.44} & \textbf{71.53} & \textbf{44.97} & \textbf{51.91} & \textbf{54.86} \\
\hline\hline
\multicolumn{9}{c}{Wiki10-31K} \\
\hline
AnnexML			& 86.46 & 74.28 & 64.20 & 77.14 & 69.44 & 11.86 & 12.75 & 13.57 \\
DiSMEC          & 84.13 & 74.72 & 65.94 & 76.96 & 70.33 & 10.60 & 12.37 & 13.61 \\
ProXML          & 85.25 & 76.53 & 67.33 & 78.66 & 71.77 & 17.17 & 16.07 & 16.38 \\
PfastreXML		& 83.57 & 68.61 & 59.10 & 72.00 & 64.54 & \textbf{19.02} & \textbf{18.34} & \textbf{18.43} \\
Parabel      	& 84.19 & 72.46 & 63.37 & 75.22 & 68.22 & 11.69 & 12.47 & 13.14 \\
Bonsai  		& 84.52 & 73.76 & 64.69 & 76.27 & 69.37 & 11.85 & 13.44 & 14.75 \\
XML-CNN         & 81.42 & 66.23 & 56.11 & 69.78 & 61.83 & \ \,9.39 & 10.00 & 10.20 \\
AttentionXML-1	& \it{87.05} & \it{77.78} & \it{68.78} & \it{79.94} & \it{73.19} & \it{16.20} & \it{17.05} & \it{17.93} \\
AttentionXML    & \textbf{87.47} & \textbf{78.48} & \textbf{69.37} & \textbf{80.61} & \textbf{73.79} & 15.57 & 16.80 & 17.82 \\
\hline\hline
\multicolumn{9}{c}{AmazonCat-13k} \\
\hline
AnnexML 		& 93.54 & 78.36 & 63.30 & 87.29 & 85.10 & 51.02 & 65.57 & 70.13 \\
DiSMEC 			& 93.81 & 79.08 & 64.06 & 87.85 & 85.83 & 51.41 & 61.02 & 65.86 \\
ProXML          & 89.28 & 74.53 & 60.07 & 82.83 & 80.75 & 61.92 & 66.93 & 68.36 \\
PfastreXML		& 91.75 & 77.97 & 63.68 & 86.48 & 84.96 & \textbf{69.52} & \textbf{73.22} & 75.48 \\
Parabel      	& 93.02 & 79.14 & 64.51 & 87.70 & 85.98 & 50.92 & 64.00 & 72.10 \\
Bonsai  		& 92.98 & 79.13 & 64.46 & 87.68 & 85.92 & 51.30 & 64.60 & 72.48 \\
XML-CNN         & 93.26 & 77.06 & 61.40 & 86.20 & 83.43 & 52.42 & 62.83 & 67.10 \\
AttentionXML-1	& \it{95.65} & \it{81.93} & \it{66.90} & \it{90.71} & \it{89.01} & 53.52 & \it{68.73} & \it{76.26} \\
AttentionXML    & \textbf{95.92} & \textbf{82.41} & \textbf{67.31} & \textbf{91.17} & \textbf{89.48} & \it{53.76} & 68.72 & \textbf{76.38} \\
\hline\hline
\multicolumn{9}{c}{Amazon-670K} \\
\hline
AnnexML			& 42.09 & 36.61 & 32.75 & 38.78 & 36.79 & 21.46 & 24.67 & 27.53 \\
DiSMEC			& 44.78 & 39.72 & 36.17 & 42.14 & 40.58 & 26.26 & 30.14 & 33.89 \\
ProXML          & 43.37 & 38.58 & 35.14 & 40.93 & 39.45 & \textbf{30.31} & 32.31 & 34.43 \\
PfastreXML		& 39.46 & 35.81 & 33.05 & 37.78 & 36.69 & 29.30 & 30.80 & 32.43 \\
Parabel     	& 44.91 & 39.77 & 35.98 & 42.11 & 40.33 & 26.36 & 29.95 & 33.17 \\
Bonsai  		& 45.58 & 40.39 & 36.60 & 42.79 & 41.05 & 27.08 & 30.79 & 34.11 \\
XML-CNN 		& 33.41 & 30.00 & 27.42 & 31.78 & 30.67 & 17.43 & 21.66 & 24.42 \\
AttentionXML-1	& \it{45.66} & \it{40.67} & \it{36.94} & \it{43.04} & \it{41.35} & 29.30 & \it{32.36} & \it{35.12} \\
AttentionXML    & \textbf{47.58} & \textbf{42.61} & \textbf{38.92} & \textbf{45.07} & \textbf{43.50} & \it{30.29} & \textbf{33.85} & \textbf{37.13} \\
\hline\hline
\multicolumn{9}{c}{Wiki-500K} \\
\hline
AnnexML			& 64.22 & 43.12 & 32.76 & 54.30 & 52.23 & 23.98 & 28.31 & 31.35 \\
DiSMEC          & 70.21 & 50.57 & 39.68 & 61.77 & 60.01 & 27.42 & 32.95 & 36.95 \\
PfastreXML		& 56.25 & 37.32 & 28.16 & 47.14 & 45.05 & \textbf{32.02} & 29.75 & 30.19 \\
Parabel     	& 68.70 & 49.57 & 38.64 & 60.57 & 58.63 & 26.88 & 31.96 & 35.26 \\
Bonsai  		& 69.26 & 49.80 & 38.83 & 60.99 & 59.16 & 27.46 & 32.25 & 35.48 \\
AttentionXML-1	& \it{75.07} & \it{56.49} & \it{44.41} & \it{67.81} & \it{65.77} & 30.05 & 37.31 & 41.74 \\
AttentionXML    & \textbf{76.95} & \textbf{58.42} & \textbf{46.14} & \textbf{70.04} & \textbf{68.23} & \it{30.85} & \textbf{39.23} & \textbf{44.34} \\
\hline\hline
\multicolumn{9}{c}{Amazon-3M} \\
\hline
AnnexML 		& \it{49.30} & 45.55 & 43.11 & 46.79 & 45.27 & 11.69 & 14.07 & 15.98 \\
PfastreXML		& 43.83 & 41.81 & 40.09 & 42.68 & 41.75 & \textbf{21.38} & \textbf{23.22} & \textbf{24.52} \\
Parabel        	& 47.42 & 44.66 & 42.55 & 45.73 & 44.54 & 12.80 & 15.50 & 17.55 \\
Bonsai  		& 48.45 & 45.65 & 43.49 & 46.78 & 45.59 & 13.79 & 16.71 & 18.87 \\
AttentionXML-1	& 49.08 & \it{46.04} & \it{43.88} & \it{47.17} & \it{45.91} & 15.15 & 17.75 & 19.72 \\
AttentionXML    & \textbf{50.86} & \textbf{48.04} & \textbf{45.83} & \textbf{49.16} & \textbf{47.94} & \it{15.52} & \it{18.45} & \it{20.60} \\
\hline
\end{tabular}
\end{table*}

\subsection{Impact of height and maximum cluster size}
Table \ref{tab:hyper:k} shows how different maximum cluster sizes $M(=K)$ 
effect the performance of AttentionXML. For keeping the number of candidates, 
we use a corresponding $C$ for different $K$. We can see that the setting of 
a smaller $M$ achieves a better performance on all datasets, especially on "tail 
labels". Table \ref{tab:hyper:h} shows how different heights $H$ effect the 
performance of AttentionXML. As shown in Table \ref{tab:hyper:h}, a smaller $H$ 
achieves a better performance. However, a setting of a smaller $M$ and a smaller
$H$ needs more time cost. So choosing these hyper-parameters is a trade-off 
between performance and time cost .

\section{Effectiveness of Attention}
We show a typical case, to demonstrate the advantage of attention mechanism in
AttentionXML. Fig. \ref{fig:attention} shows a typical text example from test 
data of Wiki10-31K. One of its true labels is ``gmail'', which is ranked at
the top by AttentionXML (while at the over 100th without multi-label 
attention). In Fig. \ref{fig:attention}, each token is highlighted by its 
attention score to this label computed by AttentionXML. We can see that 
AttentionXML gives high scores to ``Gmail'', ``e-mail'' and ``attachments'', 
which are all relevant to the true label ``gmail''. This result shows that the
attention mechanism of AttentionXML is effective for XMTC.

\begin{table*}[!t]
\centering
\caption{Performance comparisons of different $M=K$(with corresponding $C$) for 
AttentionXML.}
\label{tab:hyper:k}
\begin{tabular}{@{}ccccccccccccccccccccc@{}}
\hline
Methods & P@1=N@1 & P@3 & P@5 & N@3 & N@5 & PSP@1 & PSP@3 & PSP@5 \\
\hline\hline
\multicolumn{9}{c}{Amazon-670K, $H=2$} \\
\hline
$K=8, C=160$    & \textbf{45.74} & \textbf{40.92} & \textbf{37.12} & \textbf{43.26} & \textbf{41.53} & \textbf{29.32} & \textbf{32.50} & \textbf{35.18} \\
$K=16, C=80$    & 45.13 & 40.35 & 36.60 & 42.64 & 40.95 & 28.90 & 32.02 & 34.67 \\
$K=32, C=40$    & 44.72 & 39.98 & 36.15 & 42.29 & 40.52 & 28.80 & 31.79 & 34.22 \\
$K=64, C=20$    & 44.06 & 39.00 & 35.07 & 41.32 & 39.43 & 28.36 & 30.92 & 33.06 \\
$K=128, C=10$   & 42.96 & 37.69 & 33.51 & 39.99 & 37.85 & 27.27 & 29.46 & 31.17 \\
\hline\hline
\multicolumn{9}{c}{Wiki-500K, $H=1$} \\
\hline
$K=64, C=15$    & \textbf{75.07} & \textbf{56.49} & \textbf{44.41} & \textbf{67.81} & \textbf{65.77} & \textbf{30.47} & \textbf{37.27} & \textbf{41.69} \\
$K=128, C=8$    & 74.88 & 56.16 & 43.93 & 67.46 & 65.21 & 30.16 & 36.92 & 41.05 \\
$K=256, C=4$    & 74.26 & 55.06 & 42.30 & 66.29 & 63.39 & 30.22 & 35.87 & 38.95 \\
\hline\hline
\multicolumn{9}{c}{Amazon-3M, $H=3$} \\
\hline
$K=8, C=160$    & \textbf{49.08} & \textbf{46.04} & \textbf{43.88} & \textbf{47.17} & \textbf{45.91} & \textbf{15.15} & \textbf{17.75} & \textbf{19.72} \\
$K=16, C=80$    & 48.63 & 45.64 & 43.45 & 46.76 & 45.48 & 15.02 & 17.59 & 19.50 \\
\hline
\end{tabular}
\end{table*}

\begin{table*}[!t]
\centering
\caption{Performance comparisons of different $H$ for AttentionXML on Amazon-670K.}
\label{tab:hyper:h}
\begin{tabular}{@{}ccccccccccccccccccccc@{}}
\hline
Methods & P@1=N@1 & P@3 & P@5 & N@3 & N@5 & PSP@1 & PSP@3 & PSP@5 \\
\hline\hline
\multicolumn{9}{c}{Amazon-670K, $K=8, C=160$} \\
\hline
$H=2$ & \textbf{45.74} & \textbf{40.92} & \textbf{37.12} & \textbf{43.26} & \textbf{41.53} & \textbf{29.32} & \textbf{32.50} & \textbf{35.18} \\
$H=3$ & 45.66 & 40.67 & 36.94 & 43.04 & 41.35 & 29.30 & 32.36 & 35.12 \\
$H=4$ & 45.29 & 40.47 & 36.73 & 42.83 & 41.13 & 28.88 & 32.08 & 34.79 \\
\hline
\end{tabular}
\end{table*}

\begin{table*}[!t]
\centering
\caption{Performance comparisons ($P@5$) of AttentionXML with different $H$ on EUR-Lex, Wiki10-31K and AmazonCat-13K. $H=0$ means without a PLT.}
\label{tab:hyper:h:small}
\begin{tabular}{@{}ccccccccccccccccccccc@{}}
\hline
AttentionXML & H & EUR-Lex & Wiki10-31K & AmazonCat-13K\\
\hline
No PLT & 0 & \textbf{61.10} & \textbf{68.78} & \textbf{66.90} \\
Shallow & 2 & 60.88 & 67.27 & 66.28 \\
Deep & 4 & 60.54 & 65.89 & 65.46  \\
\hline
\end{tabular}
\end{table*}

\end{document}